\documentclass[letterpaper, 10 pt, journal, twoside]{IEEEtran}

\ifCLASSINFOpdf
\else
\fi
\usepackage{graphicx}
\usepackage{amsmath}
\usepackage{hyperref}
\usepackage{cite}

\usepackage{algorithm2e}
\RestyleAlgo{ruled}
\usepackage{amsfonts}
\usepackage[caption=false,font=normalsize,labelfont=sf,textfont=sf]{subfig}
\usepackage{stfloats}
\usepackage{accents}
\usepackage[export]{adjustbox}

\newcommand{\ubar}[1]{\underaccent{\bar}{#1}}

\usepackage[dvipsnames]{xcolor}
\usepackage[deletedmarkup=sout, authormarkup=superscript]{changes}
\definechangesauthor[name={Todo}, color=Red]{TD}

\definechangesauthor[name={Movie_todo}, color=NavyBlue]{Movie}

\definechangesauthor[name={Bram vdb}, color=OliveGreen]{bv}

\definechangesauthor[name={Gaoyuan L}, color=Cyan]{gl}

\definechangesauthor[name={Yuri Durodie}, color=Violet]{yd}

\definechangesauthor[name={Joris DW}, color=Gray]{jw}

\usepackage{listings}

\definecolor{codegreen}{rgb}{0,0.6,0}
\definecolor{codegray}{rgb}{0.5,0.5,0.5}
\definecolor{codepurple}{rgb}{0.58,0,0.82}
\definecolor{backcolour}{rgb}{0.95,0.95,0.92}

\lstdefinestyle{mystyle}{
    backgroundcolor=\color{backcolour},   
    commentstyle=\color{codegreen},
    keywordstyle=\color{magenta},
    numberstyle=\tiny\color{codegray},
    stringstyle=\color{codepurple},
    basicstyle=\ttfamily\footnotesize,
    breakatwhitespace=false,         
    breaklines=true,                 
    captionpos=b,                    
    keepspaces=true,                 
    numbers=left,                    
    numbersep=5pt,                  
    showspaces=false,                
    showstringspaces=false,
    showtabs=false,                  
    tabsize=2
}

\begin{document}
%
\title{Optimistic Reinforcement Learning-Based Skill Insertions for Task and Motion Planning}
%
%
%


\author{Gaoyuan Liu$^{1,2}$, Joris de Winter$^{1}$, Yuri Durodi\'e$^{1,2}$, Denis Steckelmacher$^{3}$\\Ann Nowe$^{3}$, and Bram Vanderborght$^{1,2}$
\thanks{Manuscript received: January, 15, 2024; Revised April, 20, 2024; Accepted May, 2, 2024.}
\thanks{This paper was recommended for publication by Editor Jens Kober upon evaluation of the Associate Editor and Reviewers' comments. This work was funded by the \textit{Flemish} Government under the program \textit{Onderzoeksprogramma Artifici\"ele Intelligentie (AI) Vlaanderen}, China Scholarship Council (CSC) and EU-project euROBIN (No.101070596).}
\thanks{$^{1}$ Authors are with Brubotics, Vrije Universiteit Brussel, Brussels, Belgium. {\tt\small gaoyuan.liu@vub.be}}%
\thanks{$^{2}$ Authors are affiliated to imec, Belgium}%
\thanks{$^{3}$ Ann Nowe and Denis Steckelmacher are with the Artificial Intelligence (AI) Lab, Vrije Universiteit Brussel, Brussels, Belgium.}%
\thanks{Digital Object Identifier (DOI): see top of this page.}
}

\markboth{IEEE Robotics and Automation Letters. Preprint Version. Accepted May, 2024}
{Liu \MakeLowercase{\textit{et al.}}: Optimistic RL Skill Insertions for TAMP} 

%



\maketitle

\begin{abstract}
Task and motion planning (TAMP) for robotics manipulation necessitates long-horizon reasoning involving versatile actions and skills. While deterministic actions can be crafted by sampling or optimizing with certain constraints, planning actions with uncertainty, i.e., probabilistic actions, remains a challenge for TAMP. On the contrary, Reinforcement Learning (RL) excels in acquiring versatile, yet short-horizon, manipulation skills that are robust with uncertainties. In this letter, we design a method that integrates RL skills into TAMP pipelines. Besides the policy, a RL skill is defined with data-driven logical components that enable the skill to be deployed by symbolic planning. A plan refinement sub-routine is designed to further tackle the inevitable effect uncertainties. In the experiments, we compare our method with baseline hierarchical planning from both TAMP and RL fields and illustrate the strength of the method. The results show that by embedding RL skills, we extend the capability of TAMP to domains with probabilistic skills, and improve the planning efficiency compared to the previous methods.
\end{abstract}


%
\IEEEpeerreviewmaketitle

\section{Introduction}

\begin{figure}
    \centering
    \includegraphics[width=0.48\textwidth]{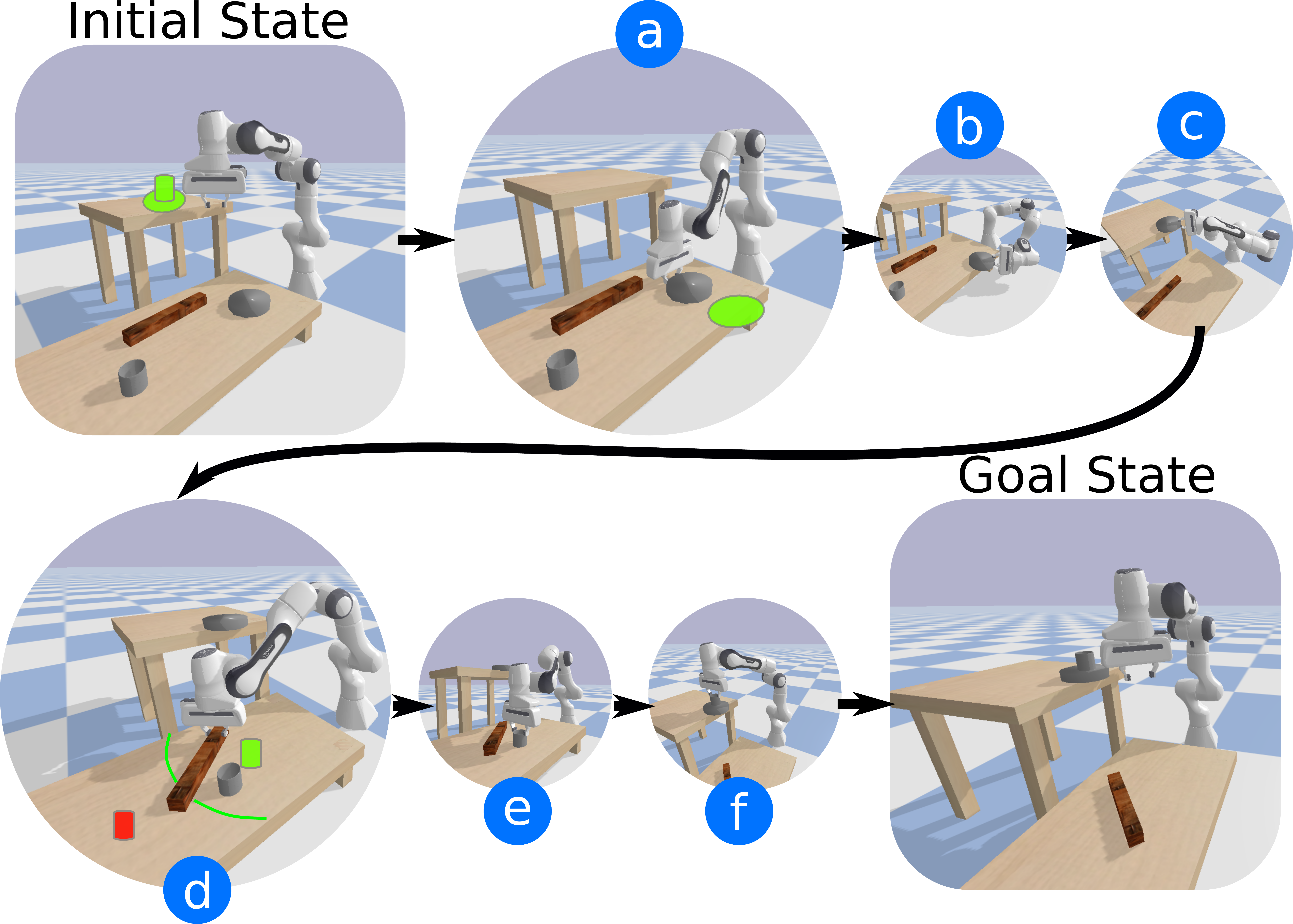}
    \caption{A table rearranging task. In this scenario, the goal is to deliver a plate and a cup on the table to another. Besides deterministic actions such as pick and place (b, c, e, f), it requires several challenging skills to achieve such a task such as (a) pushing the plate to the edge of the table in order to obtain grasping space and (d) retrieving the cup back to the workspace with the bar.}
    \label{fig:motivation}
\end{figure}

Reinforcement Learning (RL) empowers robots to acquire manipulation skills without human programming. However, prior works mostly tackle single-skill or short-term manipulation tasks, such as grasping \cite{levine2018learning} or peg insertion \cite{fan2018surreal} or synergies between two actions \cite{xu2021efficient}. The long-horizon manipulation planning remains a challenge in the RL field because of expanding state/action spaces and sparse rewards etc \cite{beyret2019dot}. Task and motion planning (TAMP) provides a general solution for long-horizon planning problems \cite{garrett2018sampling}. 
With pre-defined action knowledge, TAMP can solve deterministic tasks, e.g. sequential pick and place, within a reasonable time. However, the requirement of pre-setting precondition and effect of actions makes it challenging for TAMP algorithms to provide solutions that involve actions with effect uncertainties, which is often the case for non-prehensile actions such as pushing and sliding. Consider a scenario in Figure \ref{fig:motivation}, where the robot needs to rearrange a cup and a plate from one table to another. Besides deterministic actions such as pick and place, several other actions are required, such as: (1) If the cup is positioned beyond the robot's workspace, a \texttt{Retrieve} action is required during which the robot can manipulate a bar to slide the cup and return it into the workspace; (2) since the diameter of the plate is larger than the range of the parallel gripper, an \texttt{EdgePush} action is necessary to push the plate to the edge of the table to enable the following grasping. There are two major challenges to integrate such actions into TAMP schemes:
(1) These actions often yield uncertain effects, which impede downstream planning; (2) These actions must satisfy multiple constraints and ensure robustness against environmental noise. RL provides a learning-based approach for obtaining novel skills without explicit heuristics or modeling. Previous work synergies the capability of TAMP with a stand-alone RL policy \cite{10080986}, but such an approach becomes lagging when deploying different RL policies. 

In this work, we formulate such actions with RL policies as probabilistic \textit{skill}s. The concept of \textit{skill} is introduced in \cite{pmlr-v205-silver23a, wang2021learning} to enable learning-based actions by augmenting the logical operators with extra geometrical components.
The probabilistic skills in our work are featured as actions with bounded effect uncertainties and steered by a RL policy. The effect uncertainties are included at both planning levels. At the symbolic level, we admit the uncertainty in the effect predicates, therefore a following observing action and a plan refinement sub-routine can be deployed automatically. At the geometric level, alongside the RL policy trained with various goals and domain randomization, efforts to enhance skill robustness include the design of a data-driven state discriminator and sub-goal generator. These components are aimed at verifying predicates and providing optimistic substitutions for uncertain effects.

The remainder of the paper is organized as follows. Section \ref{sec:related_work} presents an overview of the related work. Section \ref{sec:methodology} details the methods and concepts in our framework, Section \ref{sec:experiment_and_results} analyzes the learning performance of the presented framework and demonstrates the obtained results. Section \ref{sec:conclusion} concludes the paper.

\section{Related Work}\label{sec:related_work}

The mainstream of TAMP research can be categorized into sampling-based \cite{kaelbling2011hierarchical, garrett2017sample, hauser2011randomized, garrett2020pddlstream, garrett2020online, garrett2018sampling} and optimization-based \cite{toussaint2015logic,toussaint2018differentiable, migimatsu2020object}. Garrett et al. provide an open-source framework for the sampling-based TAMP, i.e., PDDLStream, which considers deterministic actions \cite{garrett2020pddlstream}. The following work boosts the planning speed by adding geometrical intuition \cite{khodeir2023learning}. The optimization-based TAMP is usually formulated as a Logic-Geometric Programming (LGP) problem where the loss function is defined in multiple stages and their adjacent states \cite{toussaint2015logic}. As the optimization approaches tend to be slow, Driess et al. introduce a learning-based strategy that learns from the pre-planned TAMP trajectory and transforms the time-consuming planning process to a fast reaction of the deep neural network \cite{driess2020deep}. A similar manipulation function demonstrated in our work is achieved with optimization-based TAMP in \cite{migimatsu2020object}. While we try to solve the uncertainties in the motion level, they solve the probabilistic dynamics by combining an interactive controller at the low level. More learning-derived methods are designed to improve the efficiency \cite{driess2020deep}, scalability \cite{chitnis2016guided, khodeir2023learning}, and uncertainty \cite{curtis2022long} of TAMP. Moreover, RL combing TAMP is introduced to enable the robot to handle environment uncertainty \cite{yang2018peorl} and to learn the logical sequence of sub-tasks \cite{sutton1999between, blaes2019control}. 

The uncertainty has been one of the major challenges in TAMP. Kaelbling et al. categorizes uncertainties in TAMP into current-state uncertainty and future-state uncertainty \cite{kaelbling2013integrated}, and extend the previous TAMP algorithm, i.e., HPN (hierarchical planning in the now) \cite{kaelbling2011hierarchical} into the belief space. They consider the domain uncertainties while our approach focuses on the inherent uncertainties of actions. Curtis et al. solve the objects' uncertainties by leveraging computer vision techniques and predicting affordances \cite{curtis2022long}. Moreover, an assistive RL loop can be integrated into TAMP in order to improve the adaptivity of symbolic planning \cite{jiang2019task}. In this work, we focus on the action uncertainties. Instead of pre-defined action primitives, the symbolic operators and skills can be learned from examples and dataset \cite{aineto2018learning,wang2021learning,curtis2022discovering,silver2021learning,silver2022learning}. Silver et al. formalize the operator learning problem in TAMP and propose the learning operators (LOFT) algorithm in which symbolic operators (actions) are learned by aggregating the similar affect examples in the given dataset \cite{silver2021learning}. Instead of learning logical transitions, more refined skill policies and their corresponding sub-goals are learned in \cite{silver2022learning}. There are several differences between our work and \cite{silver2022learning}. Firstly, our agent learns from trial and error instead of demonstrations. Secondly, while the planning uncertainties come from inaccurate state abstraction in \cite{silver2022learning}, we focus on the inevitable effect uncertainties of actions. 
Our work is inspired by \cite{wang2021learning}, in which they introduce a comprehensive learning-based TAMP system, and the skills are comprised of a parameterized policy, an initiation set, and a termination setting. Instead of the uncertainty of the objects, our work focuses on the uncertainty of the action effect. Instead of learning the parameters of the policy with the Gaussian process, we use RL to learn policies which is essential to avoid dead-end consequences. Learning TAMP action components is also explored in \cite{mao2023learning}. However, instead of learning a parameterized sampler, we use pre-trained RL policies with domain randomization to avoid sampling parameters with multiple constraints during planning. 
Liu et al. use RL to learn non-prehensile actions which help deterministic TAMP to form a solvable situation \cite{10080986}. However, the system activates the non-prehensile action when further planning fails, therefore, the non-prehensile actions are not fully integrated in the TAMP pipelines. To amend it, our system fully integrates the RL non-prehensile actions.



\section{Methodology}\label{sec:methodology}
The TAMP pipeline we use follows a search-then-sample route. We use the Planning Domain Definition Language (PDDL) \cite{aeronautiques1998pddl} to formulate the symbolic planning domain. The essence of TAMP is defining the interaction between the symbolic (task) level and the geometric (motion) level \cite{garrett2018sampling}. That is, each action should be associated with mechanisms that can (1) verify the predicates (i.e., $\texttt{p} \in \mathcal{P}$) in the symbolic level preconditions and effects, and (2) ground the geometry level values such as the observation of the objects. In the following sections, we elaborate on how our neuro-symbolic skills are designed to handle the two-level interaction, and how the effect uncertainty is tackled in each component.

\subsection{Reinforcement Learning Skills}

\begin{figure*}
    \centering
    \vspace{2 mm}
    \includegraphics[width=0.95\textwidth]{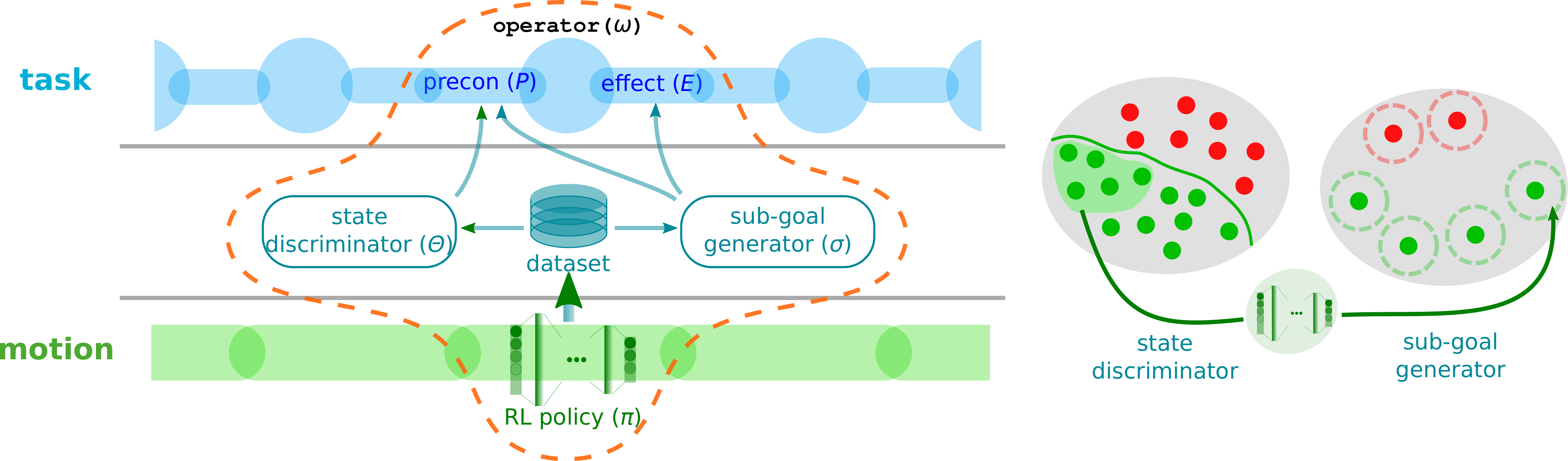}
    \caption{
    The structure of a RL skill in TAMP. The blue pipeline indicates the task planning while the green pipeline is the motion planning, they are mediated by a middle layer indicated by the turquoise entities. 
    Our RL probabilistic skills contain a RL policy in the motion level, a data-driven state discriminator, and a sub-goal generator in the middle layer. The RL probabilistic skills can connect with an operator by generating symbolic preconditions and effects. Therefore, the orange dash line frames out a RL skill.
    On the right side, the output from the state discriminator and sub-goal generator are shown. The state discriminator is a binary classifier network that can discriminate the valid initial states (in green) and invalid initial states (in red) for a skill. The sub-goal generator provides reachable sub-goals (in green) according to the initial state and the used RL policy, while unreachable sub-goals (in red) will be filtered out. The dash lines around these states represent the uncertainty boundaries of the predicted final state since the outcome of probabilistic actions, like \texttt{edgepush} or \texttt{retrieve}, cannot be determined beforehand.}
    \label{fig:structure}
    
\end{figure*}

We consider the neuro-symbolic skills as an extended version of the definition in \cite{silver2022learning}. A skill is a tuple $\phi=\langle\bar{v}, \omega, \pi, \Theta, \sigma \rangle$, it contains a tuple of object arguments $\bar{v}$; a symbolic operator $\omega=\left\langle\bar{v}, P, E\right\rangle$ where $P$ is a set of preconditions and $E$ is a set of effects; $\pi$ is a policy which contains the low-level motion plans to execute such skills; $\Theta$ is a state discriminator; and a sub-goal generator $\sigma$. The structure of a RL skill in the TAMP pipelines is shown in Figure \ref{fig:structure}. We deliberate the PDDL definition of our RL skills in Listing \ref{lst:pddl}. The definition allows uncertain effects by using the predicate \texttt{Around} instead of \texttt{AtPose}. This informs the planner to insert a subsequent \texttt{Observe} action to obtain the accurate effect state, which we further discuss in Section \ref{sec:plan_adjust}.  

\begin{lstlisting}[caption=PDDL Definition for Skills, label={lst:pddl}] 
  (:action Retrieve
    :parameters (?a ?o ?p ?x_0 ?x_g)
    :precondition (and (Arm ?a) (Pose ?o ?p)
                       (HandEmpty ?a)
                       (AtPose ?o ?p)
                       (CanRetrieveFrom ?x_0)
                       (CanRetrieveTo ?x_0 ?x_g))
    :effect (and (not (AtPose ?o ?p))
                 (Around ?o ?x_g)))
  (:action EdgePush
    :parameters (?a ?o ?p ?x_0 ?x_g) 
    :precondition (and (Arm ?a) (Pose ?o ?p)
                       (HandEmpty ?a)
                       (AtPose ?o ?p)
                       (CanPushFrom ?x_0)
                       (CanPushTo ?x_0 ?x_g))
    :effect (and (not (AtPose ?o ?p))
                 (Around ?o ?x_g)))
\end{lstlisting}

\subsection{Policy Training}
\label{sec:policy_training}
In this section, in order to illustrate that diverse policies $\pi$ (numerical-based and image-based) can be integrated into our system, we elaborate the RL training details with two example skills: \texttt{Retrieve} (action \textbf{d} in Figure \ref{fig:motivation}) and \texttt{EdgePush} (action \textbf{a} in Figure \ref{fig:motivation}). The key designs of a RL training process contain observation space, action space, and reward function. 

\subsubsection{Observation Spaces}
The observation space $x$ in the \texttt{Retrieve} environment is defined with real numbers that indicate the pose (position $\textbf{p}$ and orientation $\textbf{o}$) of objects, e.g., the current and goal poses of the bar $(\mathbf{p_b}, \mathbf{o_b})$ and the target object cup $(\mathbf{p_c}, \mathbf{o_c})$, $x = [\mathbf{p_b}, \mathbf{o_b}, \mathbf{p_c}, \mathbf{o_c}]$. For the observation space of the \texttt{EdgePush} environment, the goal of the skill is to push the object to the nearest edge. The observation should contain geometric information on both the object and the table edge, as well as their relative positions. Thus, we use the depth image as the observation. An example observation of the \texttt{EdgePush} environment is shown in Figure \ref{fig:push_train}. In this skill, we assume the object has uniform density. The uncertainties of the skill come from the contacts between the end-effector and the object, the detection error (distance and angle) of the table edge, and the shape and size of the objects. We use goal-conditioned policies, therefore the input of the policy is augmented with the assigned sub-goal, i.e., $\pi(x, x_g)$. During training, the sub-goals are uniformly sampled in the universal set of the workspace.

\subsubsection{Action Spaces}\label{sec:action_space}
The action space of both environments is a sub-space of the Cartesian space. For the \texttt{Retrieve} environment, one action is a linear motion (translation and rotation) of the bar, defined by the next pose of the bar $a = [\mathbf{p_b}', \mathbf{o_b}']$. For the \texttt{EdgePush} environment, one action is a linear motion of the end-effector which is defined by the pushing angle and distance $a = [\mathbf{\psi}_\texttt{p}, d_\texttt{p}]$. 
To simplify the problem, we assume actions are defined on a 2-D surface with a constant speed. The position of the end-effector always follows a straight line segment, and the orientation change of the end-effector is also interpolated uniformly. Afterward, the sampling-based motion planner with constraints \cite{kingston2018sampling} generates joint-space trajectories.

\subsubsection{Reward Functions}
\begin{figure}
    \centering
    \vspace{2 mm}
    \includegraphics[width=0.3\textwidth]{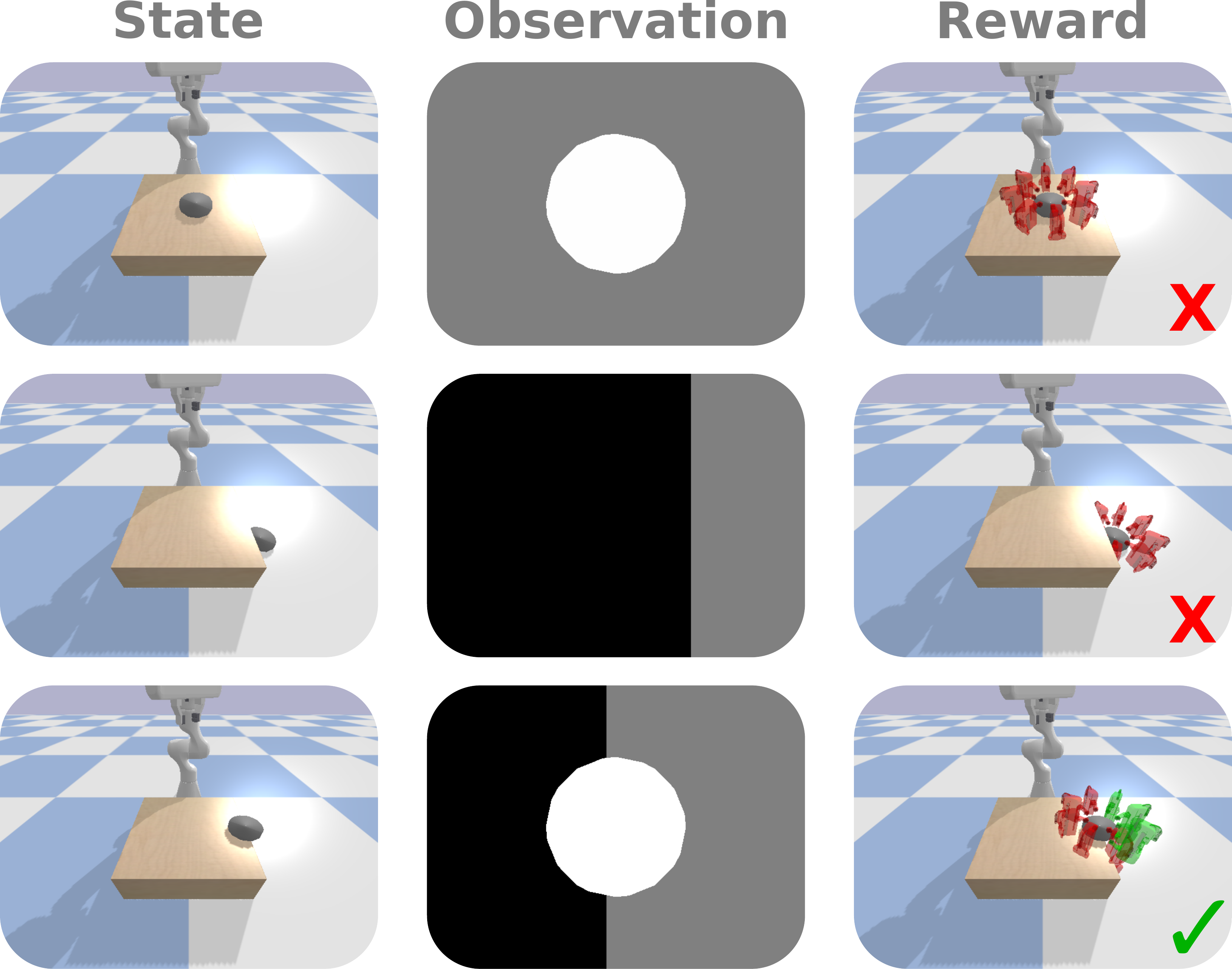}
    \caption{The \texttt{EdgePush} training environment consists of different states in the simulation presented in the first column. The second column represents the observations, i.e., the encoded depth images. According to the depth sensor, the object, table edge, and ground are encoded with different numbers. The third column illustrates that rewards will be given by sampling possible grasps. The red end-effectors indicate colliding grasping poses, while the green ones indicate collision-free grasping poses.}
    \label{fig:push_train}
\end{figure}

As for the reward, we can encode multiple considerations in one reward function. For the \texttt{Retrieve} environment, the criterion is to reach the goal position, i.e., $r = k \cdot \texttt{goal-reached}(x, x_g)$, where $g$ is defined as the goal state. For the \texttt{EdgePush} environment, the expected effect will contain multiple criteria, i.e., $r = k_1 \cdot \texttt{goal-reached}(x, x_g) + k_2 \cdot\texttt{graspable}(x) - k_3 \cdot \texttt{off-table}(x)$, where $k$s are positive factors which can be regarded as hyper-parameters. The criteria in the reward function can be conducted in the simulation. For example, as shown in Figure \ref{fig:push_train}, the $\texttt{graspable}(\cdot)$ criterion can be conducted by attempting several grasping poses. The \texttt{goal-reached} criteria is defined with a tolerance range, i.e., $\texttt{goal-reached}(x, x_g) = (||x-x_g||) \leq \epsilon$. The $\epsilon$ is the margin of the tolerable uncertainty. 

\subsubsection{Domain Randomization}
We use the domain randomization technique to improve the robustness of the policies, therefore the RL skills can tackle domain noises and help to improve the optimism of the sub-goal in Section \ref{sec:sub-goal_generator}.
More concretely, we add the following domain randomness during training: (1) The kinetic friction coefficients of objects are defined with a uniform distribution, i.e., $f_{k} = U(\bar{f}_k, \ubar{f}_k)$, where $\bar{f}_k$ and $\ubar{f}_k$ are the upper bound and the lower bound of the friction coefficient; (2) The shape of the objects in the environments vary during training, i.e., $P(\bar{v}=v| v \in \mathcal{V}) = \frac{1}{|\mathcal{V}|}$, where $\mathcal{V}$ is the set of shapes. In our work, we consider the most common shapes of table wares, i.e., $\mathcal{V} = \{\texttt{Cylinder}, \texttt{Box} \}$; (3) In each training episode, the size of objects, defined as half extent $r_{\bar{v}}$, are given with uniform distribution, i.e., $r_{\bar{v}} = U(\bar{r}_{\bar{v}}, \ubar{r}_{\bar{v}})$, where $\bar{r}_{\bar{v}}$ and $\ubar{r}_{\bar{v}})$ are the upper bound and the lower bound of the half extent; (3) In the \texttt{EdgePush} environment, we add noise on the angle of the table edge as a Gaussian $e_{\texttt{Edge}} = \mathcal{N}(\mu_{\texttt{Edge}}, \sigma_{\texttt{Edge}})$.


\subsection{State Discriminator and Sub-Goal Generator}
To integrate our RL skills in the task skeleton searching pipelines, we design the data-driven \textit{stream}-like neuro-symbolic connectors to provide state and sub-goals and their corresponding predicates. Specifically, a state discriminator to discern if the current state is in the domain of the RL skill, and a sub-goal generator to provide a sub-goal for the policy to pursue and the effect of the action. The sub-goal will also be used as the optimistic substitution during planning to guarantee the certainty assumption and will be refined later in the plan refinement process to tackle the actual uncertainty. 

\subsubsection{Data Generation}\label{sec:data_generation}
\begin{figure}
    \centering
    \includegraphics[width=0.38\textwidth]{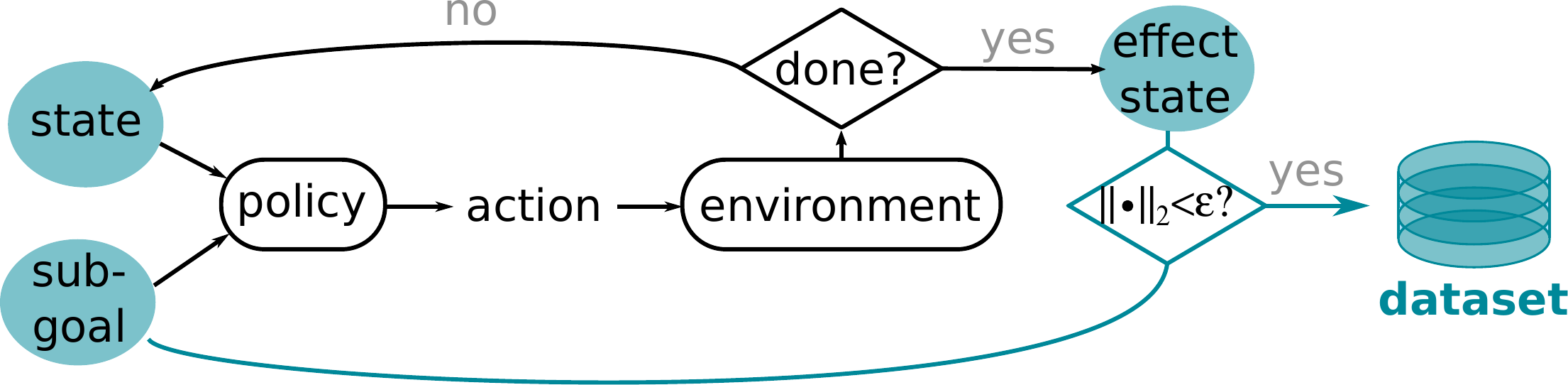}
    \caption{The data generation for state discriminator and sub-goal generator. After training, we run episodes with the trained policy. After each episode, we compare the actual final effect state and the initial sub-goal, if they are close enough then we save the initial state and the sub-goal in this episode. the black lines indicate the policy-environment interaction loop and the blue lines and shades indicate the data selection.}
    \label{fig:data_generator}
\end{figure}

As shown in Figure \ref{fig:data_generator}, we generate a dataset by running episodes in the actual planning scenario, with the trained RL policy. For each episode, an initial state and a sub-goal are randomly chosen in the universal set of the workspace.  After each episode, a success label is generated by the same reward function as during training. The success label indicates whether the episode reaches the desired effect. Therefore, one sample in the dataset can be formulated as $\{x_0, x_g, \hat{x}_{g}, \texttt{s}\}$, where $x_0\in\mathcal{X}$ is an initial state, $x_g \in \mathcal{X}$ is a goal state, $\hat{x}_{g} \in \mathcal{X}$ is an effect state, i.e., the actual final state of an episode, and $\texttt{s} = \{||x_g-\hat{x}_{g}||_2 \leq \epsilon\} \in \{0,1\}$ is a success label. We run $n$ episodes for each initial state and sub-goal pair to generate several effects states $\hat{x}_{g}$.
It's worth noting that $\hat{x}_{g}$ must either contain or be augmented with position information for downstream planning.
The motivations of this labeling are (1) even though the policy is trained, there is no guarantee that it can achieve arbitrary goals from any state, the labeled dataset can better represent the capability of the policy; (2) the successfully reached sub-goals therefore can be used as the optimistic substitutions in the primary planning process. 

\subsubsection{State Discriminator} \label{sec:state_discriminator}
\begin{figure}
    \centering
    \subfloat[]{\includegraphics[height=0.2\textwidth]{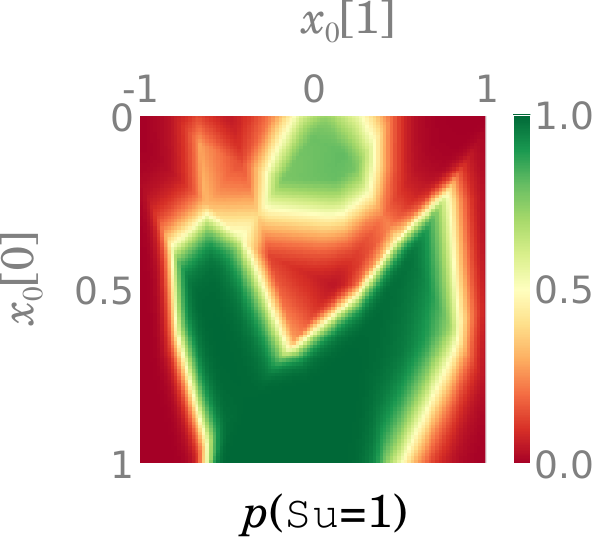}\label{fig:heatmap}} 
    \subfloat[]{\includegraphics[height=0.15\textwidth]{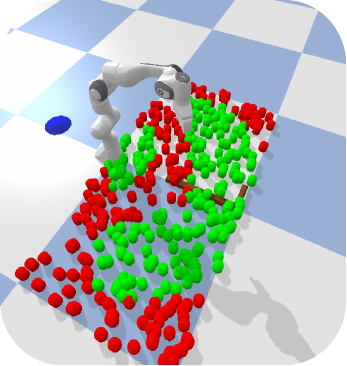}\label{fig:sd_visualization}} 
    \caption{State discriminator visualization. (a) The heatmap illustrates the predicted success possibility from the (partial) initial state $x_0$. The axes are the spatial positions. The yellow line, where $p(\texttt{s}=1)=0.5$\, is the boundary of our binary classifier. (b) We test the state discriminator by randomly sampling initial states of cups, the green cups indicate the states are valid while the red cups are invalid states.}
    \label{fig:sd}
\end{figure}

A state discriminator is a binary classifier neural network $\Theta_{\theta}$ parameterized by $\theta$. Therefore, the binary classifier satisfies:
\[
\text{Pr}(\texttt{s} = 1| \chi, \theta) = \Theta_{\theta}(\chi)
\]
Where $\text{Pr}(\cdot)$ presents the probability. The state discriminator neural network is trained by minimizing the binary cross entropy (BCE) \cite{10.5555/971143} between the target and the prediction:
\[ 
\ell(\theta)=\sum_{i=1}^m \texttt{s}^i \log \left(\Theta_{\theta}\left(\chi^i\right)\right)+\left(1-\texttt{s}^i\right) \log \left(1-\Theta_{\theta}\left(\chi^i\right)\right)
\]
Where $\chi$ is the features of the object states. To simplify the problem, here we assume the feature is the same as the initial state of each RL episode, i.e., $\chi = x_0$. It's worth noting that while more extensive features can be used to describe logical connections, this comes with the trade-off of limiting the capabilities and generalization of the policy. For skills with real-number series as observation space, such as \texttt{Retrieve} skills, we use a fully connected network, while for skills with image-based observation space, such as \texttt{EdgePush}, we use a convolutional network to better catch the features. The training processes are shown in \ref{sec:learning-skills}.
After training, the state discriminator can rule out the states that are out of the scope of the RL skills and accept states where the RL skills can be deployed confidently. In other words, it answers the question: Is the current state in the capability of the RL policy? 
During planning, the state discriminator is deployed to verify the predicates in the operator's precondition, i.e., $\mathcal{X} \times \Theta \rightarrow \mathcal{P}$. For example, $\Theta_{\texttt{Retrieve}}(x_0)=1 \rightarrow (\texttt{CanReitrieveFrom}_{p}(x_0) = \texttt{True})$, $\Theta_{\texttt{EdgePush}}(x_0)=1 \rightarrow (\texttt{CanPushFrom}_{p}(x_0) = \texttt{True})$, where the subscript $p$ indicates that the predicate is a precondition. In Figure. \ref{fig:sd_visualization}, we give a visualized example of the state discriminator of \texttt{Retrieve}. Figure. \ref{fig:heatmap} illustrates a heatmap depicting the likelihood of success from various (partial) initial states. The demarcation between the green (valid) and red (invalid) regions within the heatmap represents the critical boundary determined by the binary classifier. Figure. \ref{fig:sd_visualization} shows the effect of the state discriminator by randomly sampling the cup's states in the workspace. 



\subsubsection{Sub-Goal Generator}\label{sec:sub-goal_generator}
\begin{figure}
    \centering
    \includegraphics[width=0.42\textwidth]{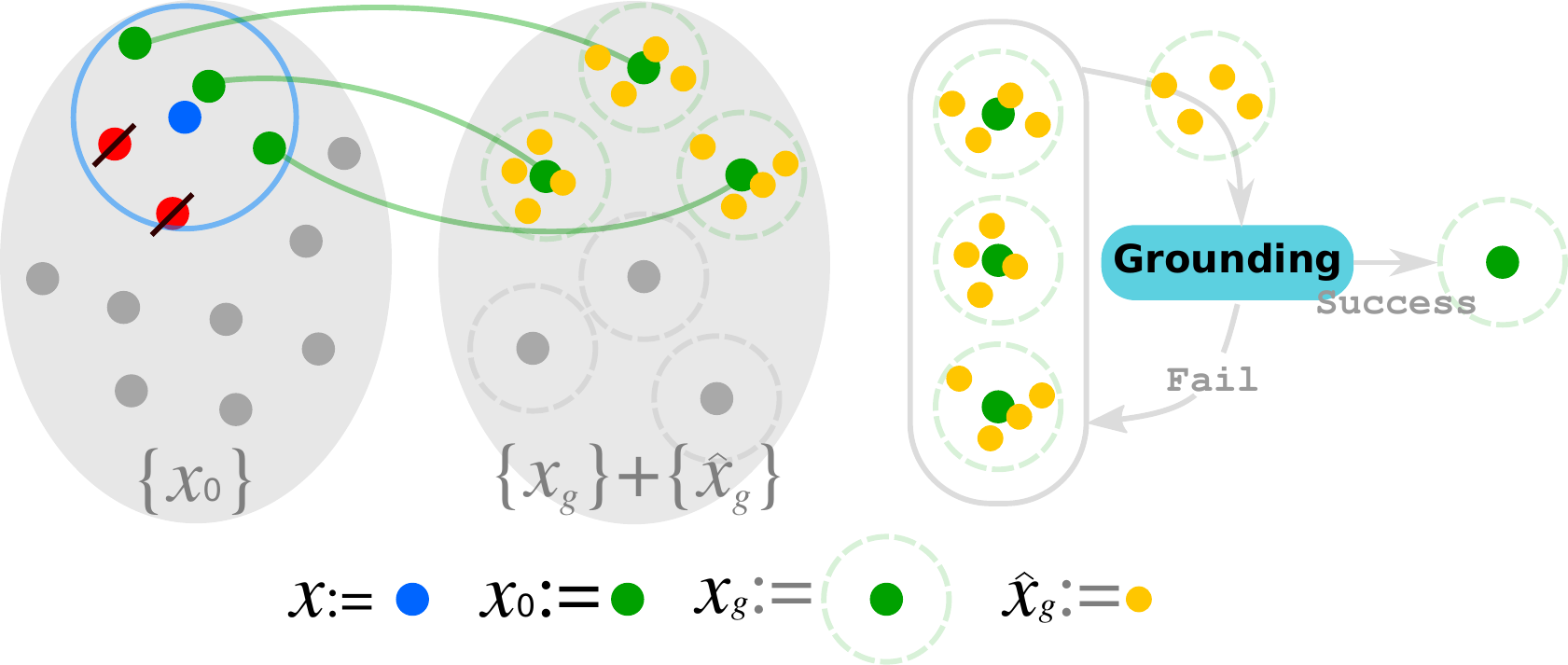}
    \caption{The sub-goal generator. Given the current state (blue dot), the sub-goal generator first finds the nearest $k$ initial states $x_0$ (green dots) in the dataset and filters out the invalid ones (red dots). The sub-goals $x_g$ (green dots with dash circles) and effect states $\{\hat{x}^{1, \cdots, n}_g\}$ that are associated with those initial states will be encapsulated as a candidate substitution sub-set. The ellipses represent sub-sets in the dataset generated in Section \ref{sec:data_generation}. The green lines indicate that the initial state $x_0$ and goal state $x_g$ are associated in the same instance. The sub-goal will be grounded during planning according to the preconditions of the subsequent action.}
    \label{fig:subgoal_generator}
\end{figure}

Because the skills have uncertain effects on the objects, we cannot specify the accurate post-action state during planning. However, the symbolic level requires the effect of actions to ground the downstream symbolic predicates and the geometric plan. To solve this challenge, we formulate the sub-goal generator as a $k$-nearest neighbor search ($k$-NN) problem, which contains the following steps: (1) Given the current state, the generator first finds the nearest $k$ initial states $\{x_0^{1, \cdots, k}\}$ and data instances that contains these initial states from the dataset; (2) exclude the data instances with the failure labels, i.e., $\texttt{s}=0$; (3) the sub-goal states $x_g$ and their associated $n$ effect states $\hat{x}_{g}$ in the rest data instances will be encapsulated and saved in a sub-set, i.e., $\{(x_{g}, \{\hat{x}^{1, \cdots, n}_g\}) | \texttt{s}=1, x_0 \in \{x_0^{1, \cdots, k}\}\}$. We define the tuple containing a goal state and its associated $n$ effect states in that sub-set as a substitution $\kappa$, i.e., $\kappa = (x_{g}, \{\hat{x}^{1, \cdots, n}_g\})$.
The candidate substitutions $\{\kappa_{1, \cdots, k}\}$ will be used in the grounding process. The workflow of the sub-goal generator is shown in Figure \ref{fig:subgoal_generator}. We define the grounded sub-goal as an optimistic substitution. 
The $k$-NN is a well-studied machine learning problem with multiple potential solutions, we choose the \textit{KDTree} method for its fast searching ability. The relevant symbolic predicates within the skill will be verified once at least one candidate substitution is found, i.e., $\mathcal{X} \times \sigma \rightarrow \mathcal{P}$. For example, $\exists \kappa \in \sigma_\texttt{Retrieve}(x) \rightarrow (\texttt{CanRetrieveTo}_{p}(x_0, x_g)=\texttt{True})\And (\texttt{Around}_{e}(x_g)=\texttt{True})$, $\exists \kappa \in \sigma_\texttt{EdgePush}(x)\rightarrow (\texttt{CanPushTo}_{p}(x_0, x_g)=\texttt{True})\And (\texttt{Around}_{e}(x_g)=\texttt{True})$, where the subscript $e$ indicates that the predicate is an effect.

\subsection{Uncertainty}\label{sec:uncertainty}
In this section, we discuss how the skills' effect uncertainties are addressed in our approach.
\subsubsection{Optimistic Substitution}\label{sec:optimistic_substitution}
The sub-goal generator outputs substitutions $\{\kappa_{1, \cdots, k}\}$ to enable downstream grounding. A substitution is composed of a sub-goal state $x_g$ and its associated $n$ effect states $\{\hat{x}^{1, \cdots, n}_g\}$. Instead of using the goals state, we use the effect state set to verify the predicates inductively. 
Specifically, we assume the predicate verification for an uncertain effect follows a Bernoulli (binary) distribution, with a maximum likelihood estimate, we can infer the \textit{optimism} $\eta_{\text{opt}}$ of a sub-goal as: 
\[
\eta_{\text{opt}}(x_g| x) = \text{Pr}(\texttt{p}(\pi(x,x_g)) = \texttt{True}) = \frac{\sum_{i=1}^{n}\texttt{p}(\hat{x}^{ i}_g)}{n}
\]
Therefore, given the current state $x$, the sub-goal with the highest \textit{optimism} will be grounded as an optimistic substitute, i.e., $x^{*}_{g}(x) = \text{argmax}_{x_g}\eta_{\text{opt}}(x_g|x)$. It is worth mentioning that the predicate $\texttt{p}$ verified here can be associated with the subsequent action. For example, effect substitutions of a \texttt{Retrieve} skill are normally used to ground the grasping pose in the following \texttt{Pick} action and to verify the inverse-kinematics predicate, i.e., $\texttt{IK}(x_g)$, in the precondition. 
By using this statistical grounding strategy, we incorporate the effect uncertainties into the planning.

\subsubsection{Plan Refinement}\label{sec:plan_adjust}
\begin{figure}
    \centering
    \vspace{2 mm}
    \includegraphics[width=0.4\textwidth]{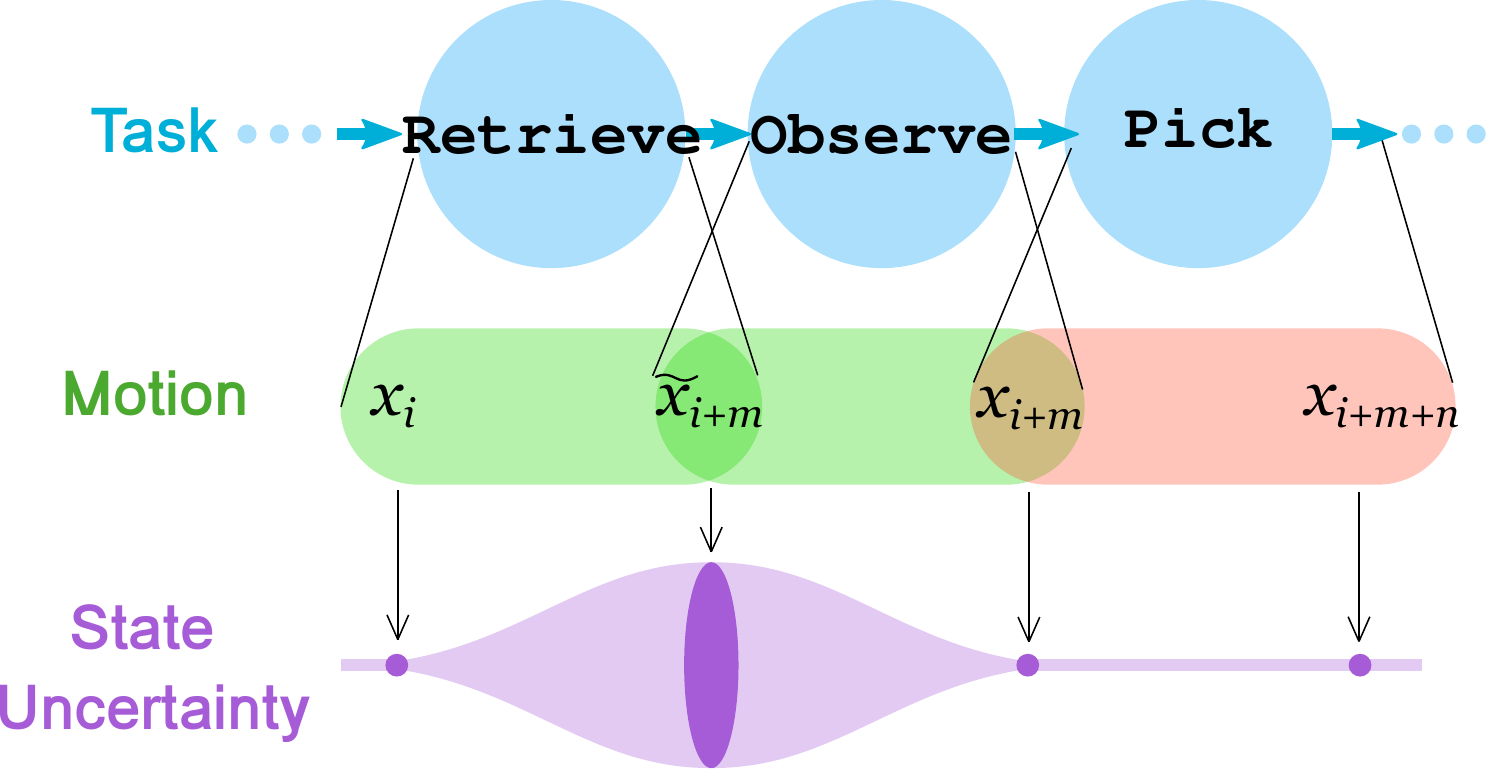}
    \caption{The plan refinement subroutine. Probabilistic skills come with inevitable uncertainties in their effect states. Such uncertainties are prone to cause failure in the following action. Two changes are necessary to mitigate the uncertainties: (1) In the task level, an \texttt{Observe} operator is inserted after each probabilistic skill; (2) In the motion level, the motion in the following deterministic task (in the red shade) will be refined based on the observed state. The purple tubes illustrate the state uncertainty, a bigger diameter indicates that the state is more uncertain.}
    \label{fig:plan_refinement}
\end{figure}

The inevitable discrepancies between the optimistic substitutions and the actual effects of probabilistic actions can lead to failure unless subsequent motions are adjusted accordingly. To solve this problem, the planning process needs to be adjusted in two ways. First, an \texttt{Observe} action is necessary after every probabilistic action. To enable the solver to deploy the \texttt{Observe} action automatically, the PDDL definitions of the skills allow the uncertain effect by using \texttt{Around} as the effect predicate (see line 9 and line 18 in the Listing \ref{lst:pddl}). Then an action that converts \texttt{Around} into \texttt{AtPose} will be deployed as \texttt{AtPose} is required in the following deterministic action's precondition. The PDDL definition of the \texttt{Observe} action is shown in Listing \ref{lst:pddl2}.
\begin{lstlisting}[caption=PDDL Definition for Observation, label={lst:pddl2}]      
  (:action Observe
    :parameters (?o ?x_g ?p)
    :precondition (and (Around ?o ?x_g)
                       (Pose ?o ?p)        
    :effect (AtPose ?o ?p))
\end{lstlisting}
Second, the motion plan of the following deterministic action should be replanned. An example is shown in Figure \ref{fig:plan_refinement}, after executing a \texttt{Retrieve} skill, the object state uncertainty increases. Therefore, the motion of the following \texttt{Pick} action should be replanned based on the result of the \texttt{Observed} action. Therefore, the execution not only imports the TAMP plan but also modifies the TAMP plan on the fly. The \texttt{observe} actions in the TAMP plan provide control signals for the following action to replan. 
It's worth noting that since the discrepancy between the real effect and the optimistic substitution is bounded, the motion of the following action should keep the feasibility of the inverse kinematics, therefore, the task-level plan will not be affected. 

\subsection{Limitations}

The proposed method has three known limitations. First, there is an intrinsic limitation regarding the generality of RL policies. While the RL policy is trained within a specific domain. We attempt to extend a skill's applicability by randomizing physics and geometry factors. Therefore, the RL policy can adapt to changes in shape, size, and friction within a certain range. 
Secondly, RL policies solely focus on the impact of objects within their observation space. 
To address this, vision-based policies can provide object number-agnostic policies.
Third, using a dataset in the system limits the scalability of our method. Further research can be conducted to address these limitations.
 

\section{Experiment and Results} \label{sec:experiment_and_results}

\subsection{Domain Setting}
\begin{figure}
    \centering
    \vspace{2 mm}
    \includegraphics[width=0.4\textwidth]{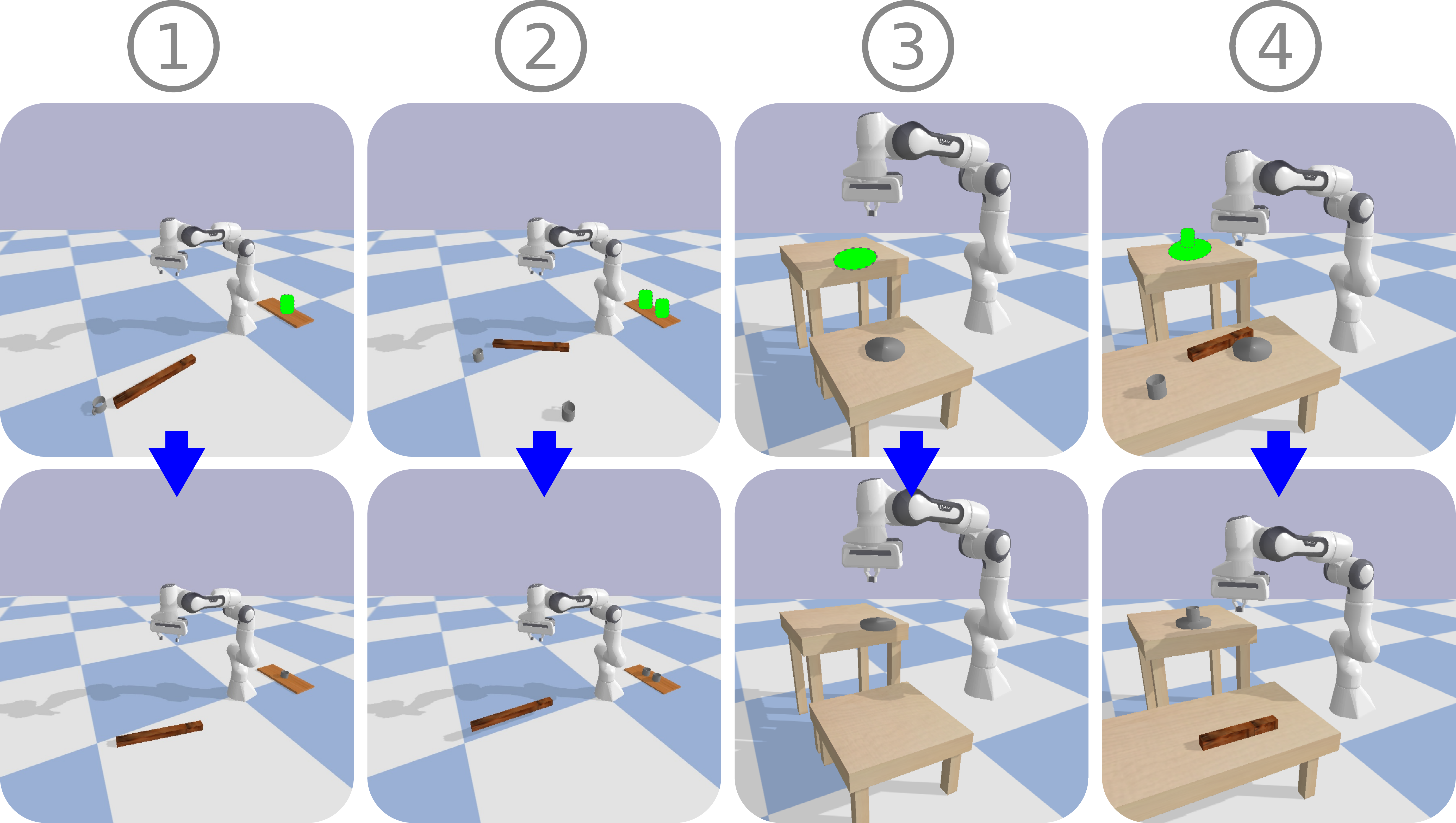}
    \caption{Experiment scenarios. We test different methods in four scenarios, from left to right: (1) Retrieving, (2) multi-retrieving, (3) edge-pushing and (4) serving. The top row illustrates the initial state of the problem in which the goal positions are highlighted in green. The bottom row is the goal state in each problem.}
    \label{fig:exp_setting}
\end{figure}

We verify our method with two kinds of probabilistic skills in four scenarios: (1) Retrieval, (2) multi-retrieving, (3) edge-pushing, and (4) serving, as shown in Figure \ref{fig:exp_setting}. The general goal of these domains is to relocate the objects in the goal positions, besides pick and place, probabilistic skills such as \texttt{Retrieve} and \texttt{push} are needed. Worth noting that different problems can illustrate different features of algorithms. For example, multi-retrieving illustrates the scalability of a single skill, and serving presents the combination of different skills. We choose the \textit{adaptive} method as the TAMP solver because it shows the best efficiency among the benchmark sampling-based TAMP solvers in \cite{garrett2020pddlstream}.

\subsection{Learning Skills}\label{sec:learning-skills}
\begin{figure}
    \centering
    \subfloat[]{\includegraphics[width=0.2\textwidth]{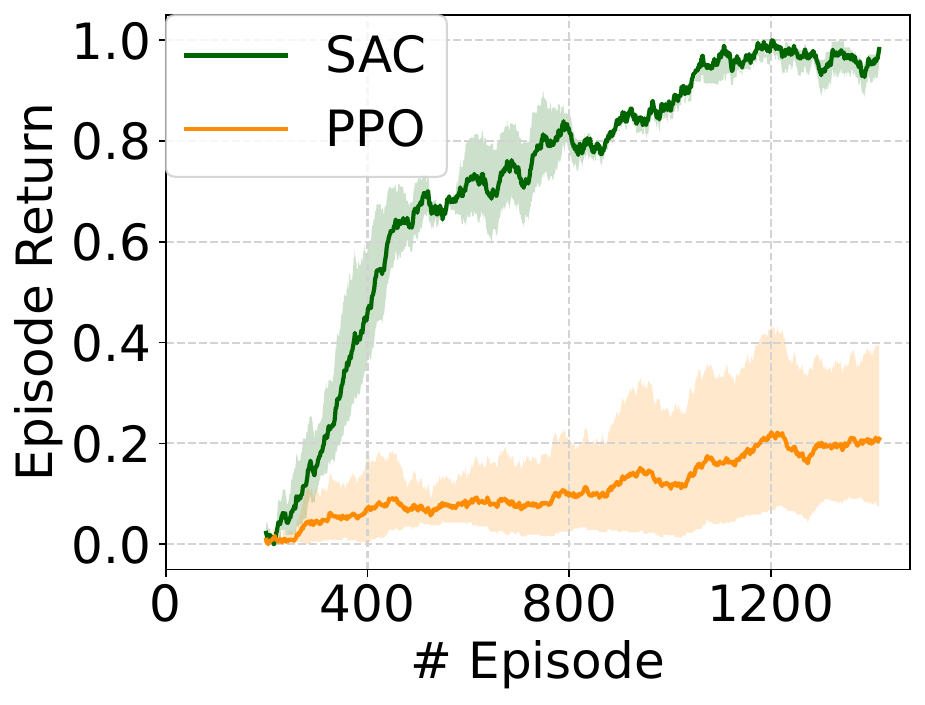}\label{fig:learning_el}} 
    \subfloat[]{\includegraphics[width=0.2\textwidth]{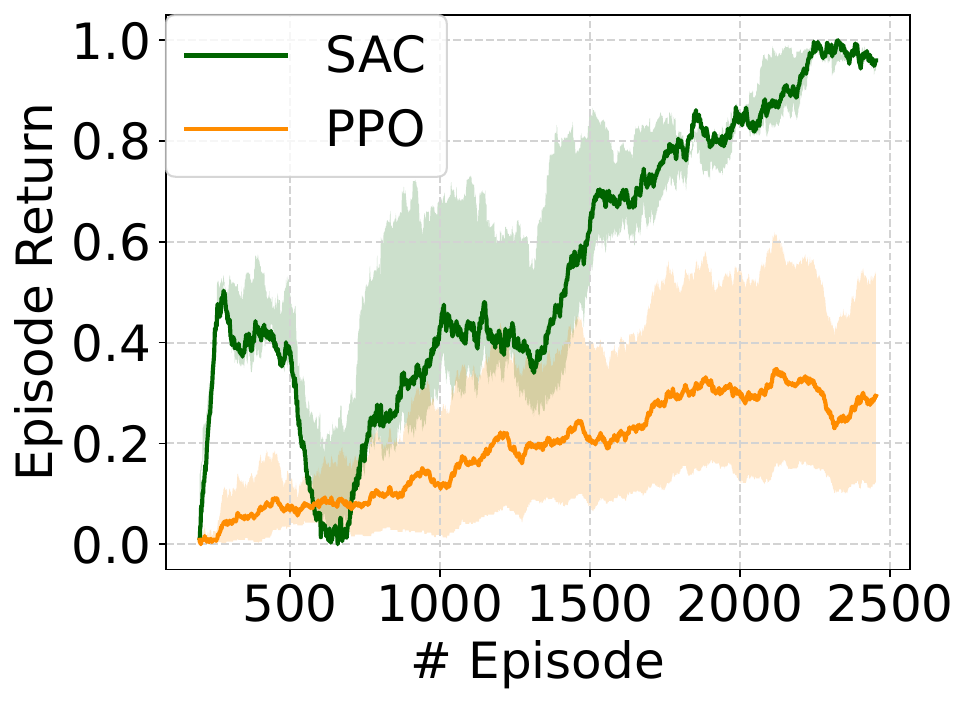}\label{fig:learning_er}} 
    \caption{The episodic accumulated reward, i.e., episode return, of the policy for (a) \texttt{Retrieve} skill and (b) \texttt{EdgePush} skill. The learning curves show that by self-learning in simulation, the agents can successfully optimize the episode return and therefore learn rational policies. The learning curves show that SAC achieves better data efficiency than PPO.}
    \label{fig:rl_learning_curves}
\end{figure}

\begin{figure}[ht]
    \centering
    \subfloat[]{\includegraphics[width=0.2\textwidth]{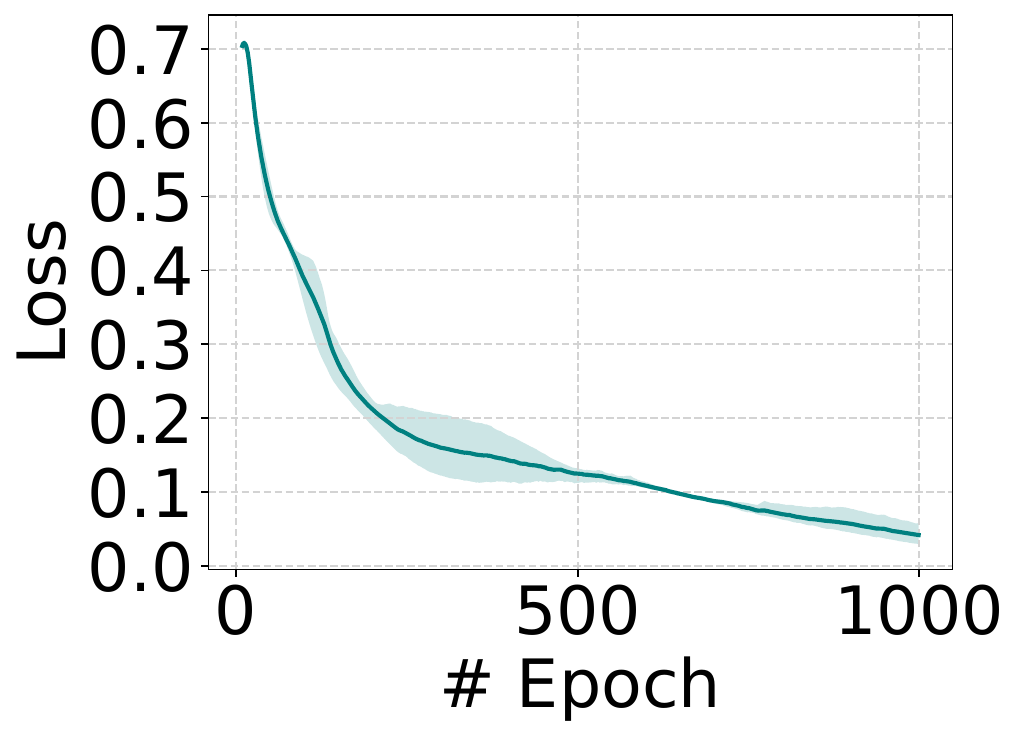}\label{fig:sd_retrieve}} 
    \subfloat[]{\includegraphics[width=0.2\textwidth]{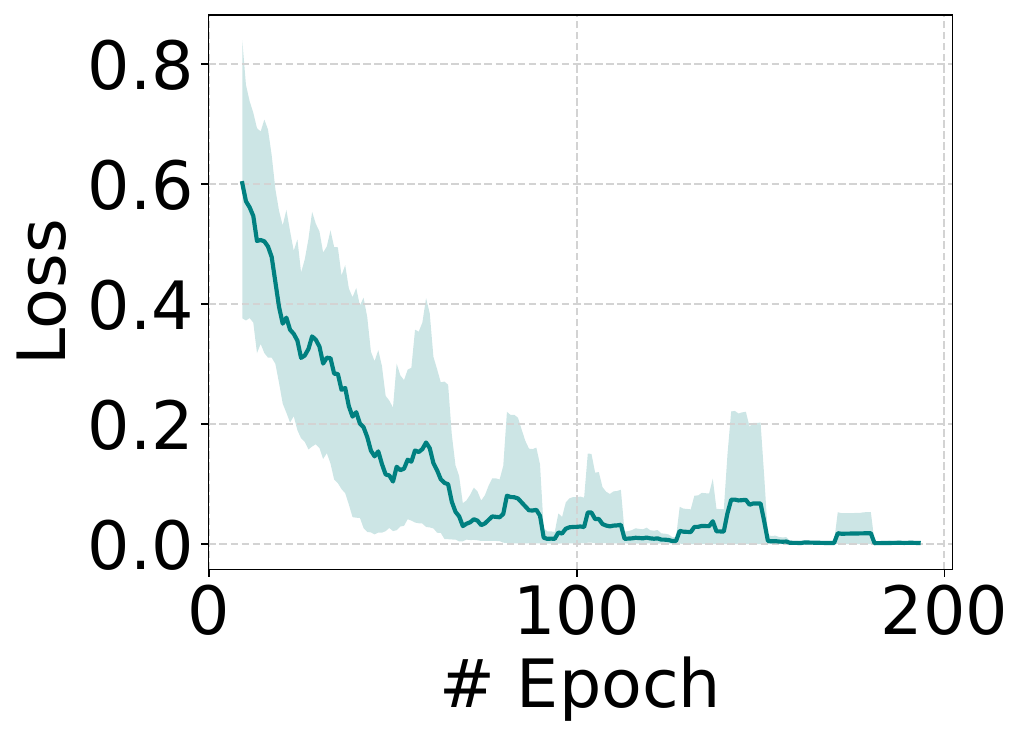}\label{fig:sd_edgepush}} 
    \caption{ 
    Training loss for state discriminators. The two figures are the training losses of the state discriminator of (a) \texttt{Retrieve} skill and (b) \texttt{EdgePush} skill. We ran each training and evaluation process five times. }
    \label{fig:sd_train}
\end{figure}

Every skill contains two neural networks: a policy network and a state discriminator network. In this section, we illustrate the training process of both networks in each example skill. The policy network follows the normal RL training courtesy and accumulates rewards in each episode. We compare the benchmark on-policy RL algorithm Proximal Policy Optimization (PPO) and off-policy Soft Actor Critic (SAC). The Learning curves are shown in Figure \ref{fig:rl_learning_curves}. 
The state discriminator network updates the weights by simply minimizing the loss between the prediction and target as discussed in Section \ref{sec:state_discriminator}. The \texttt{Retrieve} state discriminator uses a fully connected network while the \texttt{EdgePush} uses a convolutional network for the image-based observation space. The training process of state discriminators is illustrated by the loss, shown in Figure \ref{fig:sd_train}.

\subsection{Comparison Experiment}
\begin{figure}
    \centering
    \subfloat[]{\includegraphics[width=0.4\textwidth]{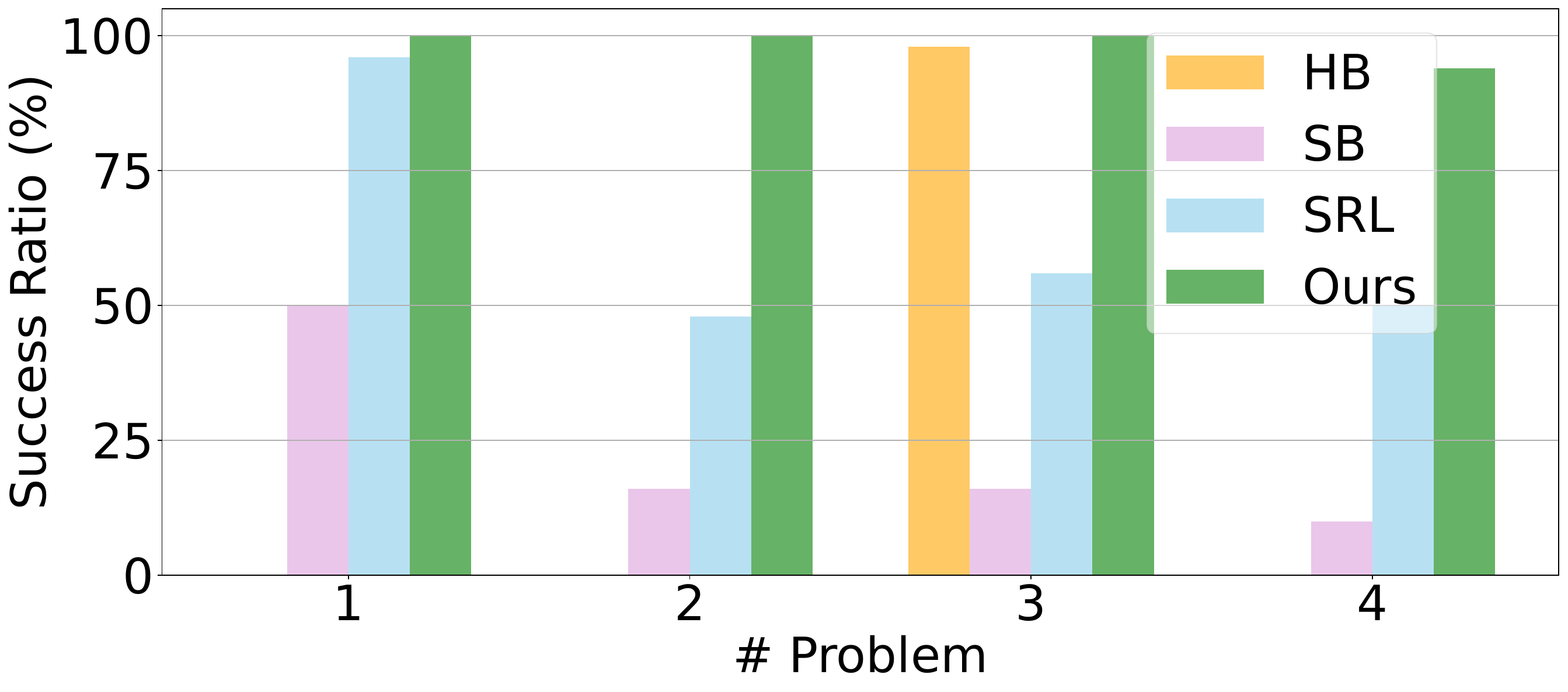}\label{fig:planning_success_ratio}} 
    \hfill
    \subfloat[]{\includegraphics[width=0.4\textwidth]{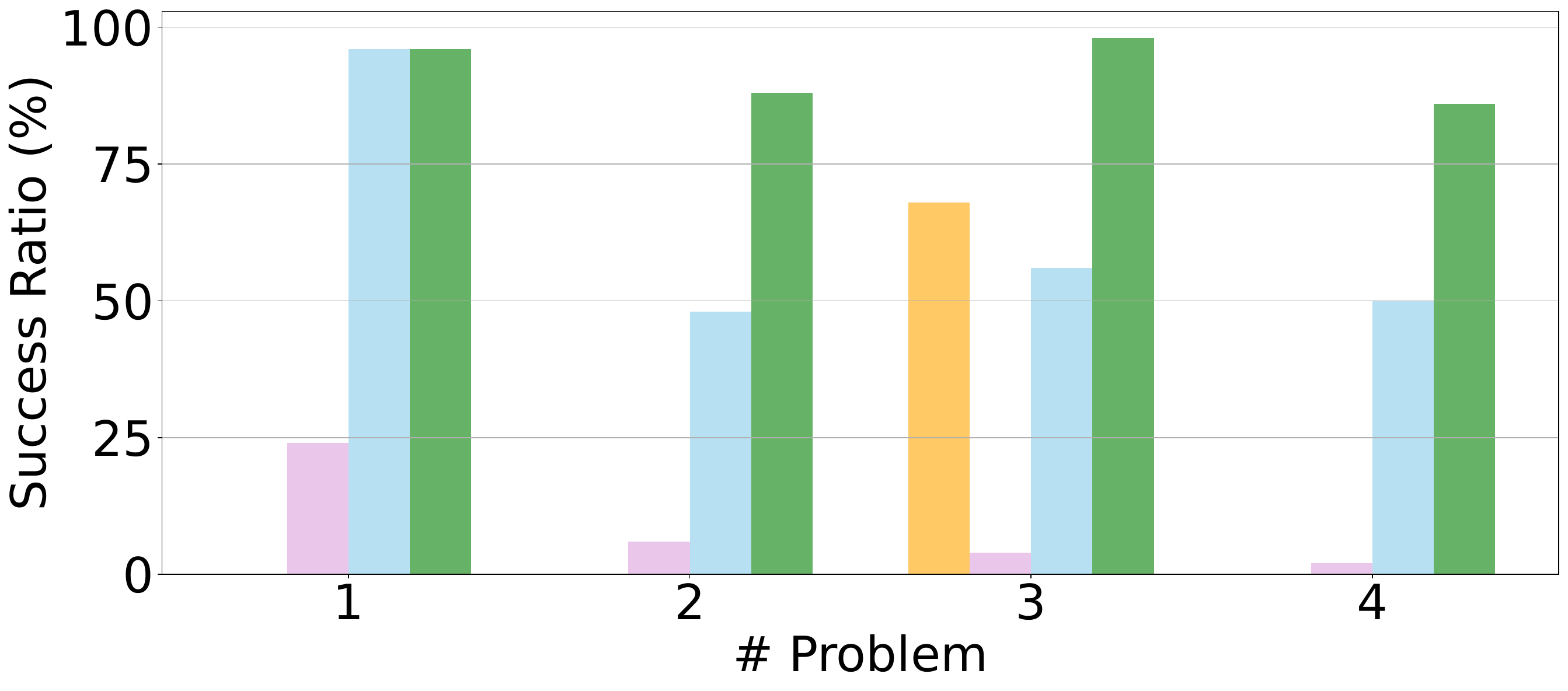}
    \label{fig:execution_success_ratio}} 
    \caption{Success ratio comparison. We compare (a) planning success ratio and (b) execution success ratio of different methods in different problems. Each color of the bars corresponds to one method, the absent bars indicate the corresponding method cannot solve the problem.}
    \label{fig:baseline_compare_success}
\end{figure}

\begin{figure}
    \centering
    \vspace{2 mm}
    \includegraphics[width=0.4\textwidth]{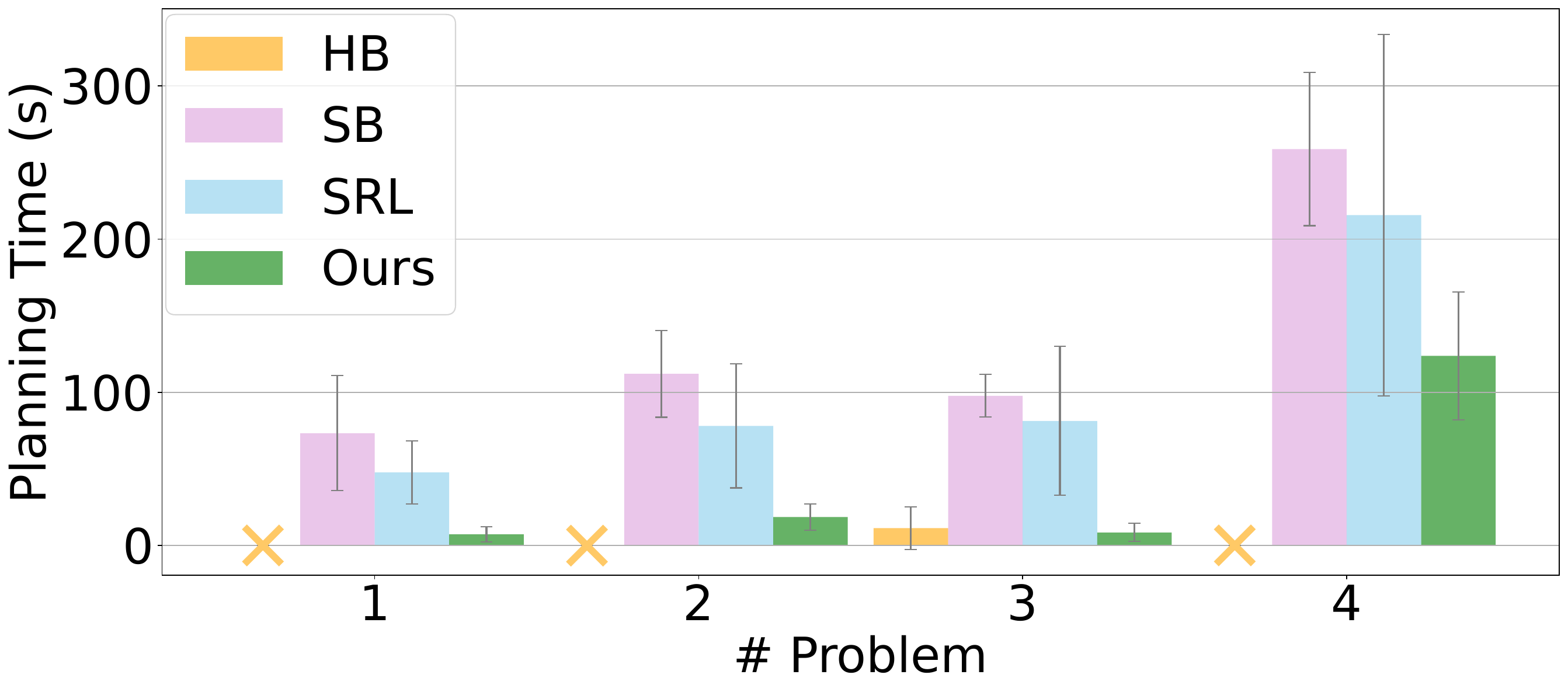}
    \caption{Planning time comparison. We compare the average planning time of different methods in different problems. The grey line segment presents the standard deviation. Each color corresponds to one method, the absent bars with crosses indicate the corresponding method is not able to solve the problem.}
    \label{fig:baseline_compare_time}
\end{figure}

We compare our system with three baseline methods: (1) Heuristic-based (HB) method, as used in \cite{garrett2015backward, garrett2018sampling}, assumes that all actions are deterministic and yield accurate effects. Consequently, no plan refinement subroutine is employed; (2) plain sampling-based (SB) solver, in which all actions are grounded by uniform sampling from the action space defined in Section \ref{sec:action_space}; (3) synergistic RL (SRL), in which the RL actions assist the sampling-based TAMP solver as a stand-alone module \cite{10080986}. We run each problem 50 times and evaluate different methods by comparing (1) planning time (in Figure \ref{fig:baseline_compare_time}), (2) planning success ratio (in Figure \ref{fig:planning_success_ratio}) and (3) execution success ratio (in Figure \ref{fig:execution_success_ratio}) of each method. The tolerant solving time is set to $150 s$ in testing scenarios except $350 s$ in the last scenario because of the higher complexity. Planning trials that exceed the tolerant solving time are considered failed. The success of execution is evaluated using the symbolic-level predicates of the final goal.
As shown in the figures, the HB and SB methods generally struggle with problems that require probabilistic skills because: First, for some skills such as \texttt{Retrieve}, there is no obvious heuristics to guide the sampling process, therefore the HB method cannot solve the problems that require \texttt{Retrieve} skills (i.e., problem 1,2 and 4). Second, sampling a probabilistic skill with forward simulation is time-consuming due to multiple motions and delayed revelation of each sampling's effect. Third, the success ratio of both HB and SB tends to drop during execution because they assume that the effect of the actions will be the same in planning and execution. The SRL can achieve high success ratios in all problems but it needs longer planning time. It's worth mentioning that the SRL method triggers all RL skills without discrimination, irrespective of the cause behind the failure of the sampling-based TAMP.
On the contrary, our method integrates RL skills into the planning process with reasoning. Therefore, our method shows better scalability comparing the performance in the first and the second problem. 
However, the extension of the problem horizon still poses a limitation for the current work, as the additional skills inevitably expand the search space. The discrepancy between success ratios of planning and execution can be caused by the residual errors between groundings and actual state.

\subsection{Real-World Experiments}
\begin{figure}
    \centering
    \vspace{2 mm}
    \includegraphics[width=0.45\textwidth]{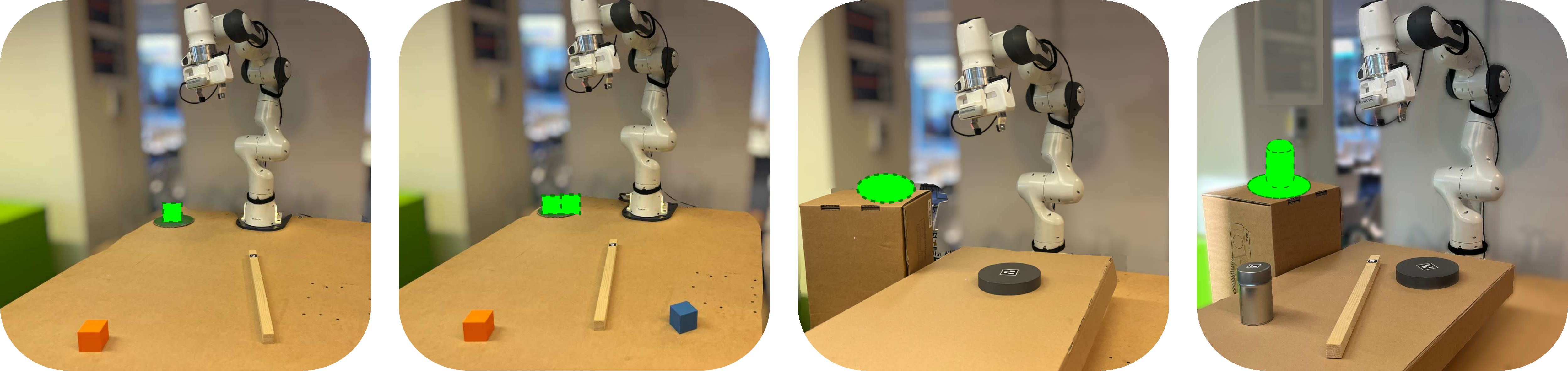}
    \caption{Real-world settings. We conduct the corresponding experiments in the real-world environment. The manipulation goals are illustrated with green shapes. }
    \label{fig:real_scn}
\end{figure}

\begin{figure}
    \centering
    \includegraphics[width=0.4\textwidth]{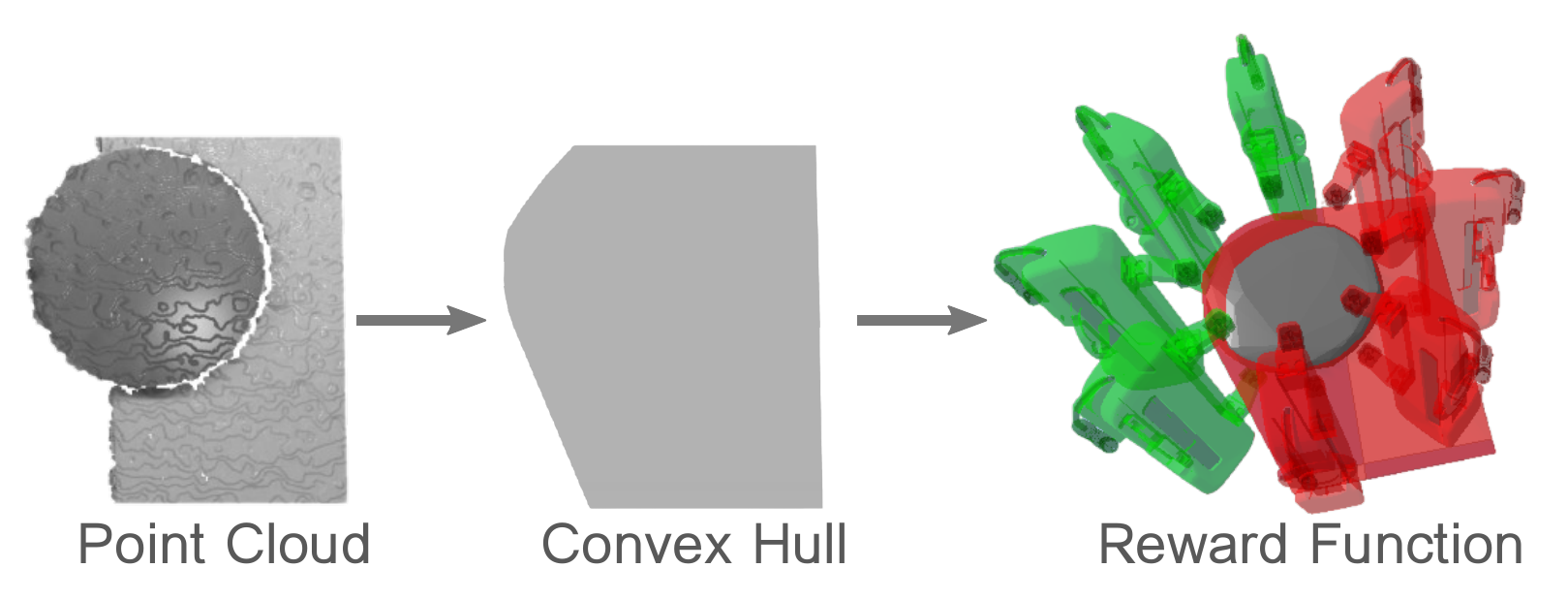}
    \caption{Detection flow in real experiments. The model of the object is unknown to the RL agent, we need to reconstruct the model and the obstacle information for the policy to terminate. We use a 3D reconstruction workflow. The camera captures depth data and converts it into a point cloud. Then a convex hull is built according to the point cloud. Finally, the convex hull of both object and obstacle can be fed back into the simulation to conduct the collision checking. }
    \label{fig:real_reward_push}
\end{figure}

Finally, we implement the proposed method in a real-world system. Specifically, Franka Emika is used as the manipulator, while an in-hand Intel RealSense D435 camera as the observer. The real-world settings are shown in Figure \ref{fig:real_scn}. The skill \texttt{EdgePush} is model-agnostic. That is, the shape of the object and the spatial relation with the table edge are unknown beforehand but can be determined by point cloud detection. The detection flow is shown in Figure \ref{fig:real_reward_push}. 
The model-agnostic skills allow the skill to function on different objects and can handle small disturbances. In the experiment, we also tested such robustness by giving the table edge small changes. The code of the proposed method and more experiment videos can be found on GitHub: \url{https://github.com/Gaoyuan-Liu/ORL-TAMP}. 



\section{Conclusion}\label{sec:conclusion}
Our main goal in this paper is to design a method that allows RL skills to be integrated into a TAMP process, thereby extending the capability of the planning scheme into domains with probabilistic skills. To achieve this goal, we encapsulate each RL policy with a state discriminator and a goal generator into a neural-symbolic skill. Such a skill can be planned using AI planning languages and solvers such as PDDL and \textit{Fast Downward}. The experiments show that our method can achieve better planning efficiency compared to the baseline sampling-based and RL-TAMP methods. Besides the limitations discussed in the paper, future research can be conducted to improve the method's versatility and scalability.



\medskip

\bibliographystyle{ieeetr}
\bibliography{reference}

\end{document}